# Qualitative Projection Using Deep Neural Networks

Andrew J.R. Simpson [#1]

[#] *Centre for Vision, Speech and Signal Processing, University of Surrey*
*Surrey, UK*
[1] `Andrew.Simpson@Surrey.ac.uk`

*Abstract*—**Deep neural networks (DNN) abstract by demodulating the output of linear filters. In this article, we refine this definition of abstraction to show that the inputs of a DNN are abstracted *with respect to* the filters. Or, to restate, the abstraction is *qualified by* the filters. This leads us to introduce the notion of *qualitative projection*. We use qualitative projection to abstract MNIST hand-written digits *with respect to* the various dogs, horses, planes and cars of the CIFAR dataset. We then classify the MNIST digits according to the magnitude of their dog*ness*, horse*ness*, plane*ness* and car*ness* qualities, illustrating the generality of qualitative projection.**

*Index terms*—Deep learning, parallel dither, unsupervised learning, qualitative projection, Yin and Yang, DSP Interpretation.

## I. INTRODUCTION

*Depth of learning* in deep neural networks (DNN) comes from the process of abstraction [1-5]; DNN are comprised of linear filters (i.e., weights) followed by nonlinear activation functions. The linear filters are selective of *features*. These selective filters output *variance* of the feature *with respect to* the input. The nonlinear activation functions are able to demodulate this *feature variance* via a process of rectification [1,2,5].

This leads to a more rigorous definition of abstraction: in a DNN, the input is *abstracted with respect to* the filter. This may be restated yet more rigorously as: *the abstraction of the input is qualified by the filter*. This *qualification* is validated by the inverse process of *synthesis*, whereby the *qualities* are *magnified* (multiplied with a magnitude). Thus, we may characterise the process of abstraction as a process of *qualitative projection*.

In this article, we use a process of *qualitative projection* to obtain an abstract similarity measure. We first train a *projector DNN* to abstract a magnitude of 1 when input with a specific *projection image*. Thus, we obtain the filters whose demodulated output best capture the qualities of the input *projection image*. We then use this *projector DNN* to abstract new images with respect to the qualities of the original *projection image*.

This *qualitative projection* provides us with an abstract similarity measure – a magnitude (1 being identical and <1 indicating less similar) capturing the abstract similarity of the input with respect to the qualities of the *projection image*. We then instantiate 100 *projector DNNs*, each trained to abstract with respect to the qualities of a different training image, and then we project an entire training and test set through the respective *projector DNNs*.

By projecting a training set of images through 100 *projector DNNs*, we obtain a similarity measure for each of the training images *with respect to* the qualities of the original *projection images*. This provides a vector representing each training example in terms of similarity to the original 100 projection images – a multi-dimensional *qualitative projection*. Thus, we describe a generalised principle of *qualitative projection* that seems compatible with the concept of 'embedding' [6,7].

We then validate this multi-dimensional *qualitative projection* by training a *classifier DNN* on the projections (rather than on the native images). During the process of training the *projector DNNs*, we show that demodulation is a pre-requisite of *qualitative projection* by illustrating the importance of both rectification via the biased-sigmoid activation function and suppression of 'decoy-features' [5] via parallel dither [3,4].

## II. METHOD

Our *qualitative projection* paradigm features CIFAR10 [8] images (e.g., dogs, horses, cars, ships, planes, etc) and MNIST [9] hand-written digits (i.e., hand-written characters from 0 to 9). Examples of both classes of image are given in Fig. 1. The idea here is to use the CIFAR images to obtain *qualitative projections* of the MNIST images. I.e., given some arbitrary (CIFAR) images we obtain a qualitative projection of each MNIST image which captures the abstract similarity between the hand-written digit and each of the CIFAR objects. For example, given an MNIST digit image, we might obtain a respective similarity vector comprising: [0.4, 0.2, 0.1, 0.6] for CIFAR images of [horse, cat, plane, ship] respectively. This corresponds to magnitudes for qualities of 0.4 horse*ness*, 0.2 cat*ness*, 0.1 plane*ness* and 0.6 ship*ness*. In other words, we obtain an abstraction vector where each element is *qualified by* the respective (arbitrary) CIFAR image. A general schematic diagram is given in Fig. 2.

*Qualitative projection*. Using 100 randomly selected images from the CIFAR10 dataset [8], 100 respective *projector DNNs* were obtained. The projector DNNs were of size 784x100x1 units (784 = 28x28 input layer, 1 = single neuron output layer). In one instantiation, typical sigmoid activation functions were used at each layer. In an alternative instantiation, biased-sigmoid [2] activation functions were used in each layer. The biased-sigmoid activation function is optimised for demodulation [2] and this may be interpreted as a form of regularisation which enforces demodulation. Hence, we hypothesise here that we can regularise for *qualitative projection* with a biased sigmoid. To test this, we include the typical sigmoid for comparison. Note that the typical sigmoid may learn to demodulate (by learning a positive bias) but, by the path of least resistance, it may also (more easily) learn arbitrary mappings (as opposed to abstractions). Hence, the biased sigmoid regularises towards abstraction (and against arbitrary learning) (see [2,5]).

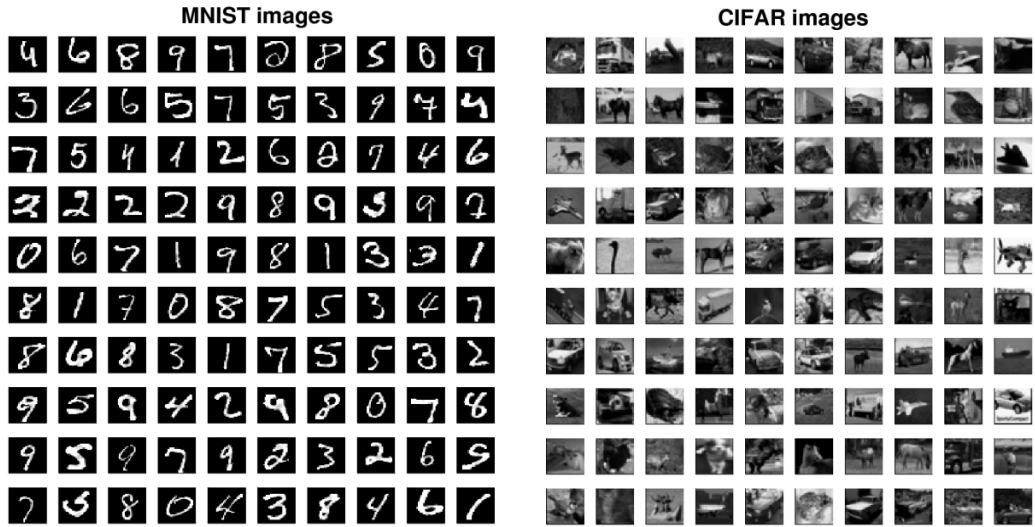

**Fig. 1. MNIST digits and CIFAR10 images.** We took the 28x28 pixel MNIST [9] images and unpacked them into a vector of length 784. We also took the 32x32 pixel CIFAR10 [8] images and resampled them to the same 28x28 scale as the MNIST images and unpacked them similarly. The CIFAR images were used to train the *projector DNN*s. Then, the *projector DNN*s were used to obtain a multidimensional *qualitative projection* of the MNIST training and test images (into a vector of dimension 100).

Each *projector DNN* was trained such that the respective CIFAR image was input and back propagation gradient descent was used to optimise the output activation towards a '1'. Thus, the *projector DNN*s learned the filters which best abstracted to a magnitude of 1 for a single image. (i.e., only one image was used to train each *projector DNN* to abstract a magnitude of 1). Each projector DNN was trained (from the typical random starting weights) for 50 iterations of gradient descent with a learning rate of 1.

*Parallel dither*. In addition to regularising for abstraction via the biased-sigmoid, we also include *parallel dither* [3-5] for the same purpose. In brief, *parallel dither* acts to suppress 'decoy features' (nonlinear products of the activation function) [5] which might otherwise result in arbitrary mapping (rather than abstract learning) via the path of least resistance. Thus, we also include an instantiation of the *qualitative projection* paradigm whereby each projector DNN was trained with 100x parallel dither and where the projections of each projector DNN (i.e., of the training and test MNIST images) was also subject to 100x parallel dither (see [5]).

*Parallel-dithered projections*. We also used *parallel dither* to suppress decoy features in the qualitative projections. Each MNIST image was replicated 100x and to each dither noise was added (uniform noise of zero mean and unit scale). These 100 dithered images were then projected using a given *projector DNN*. The 100 respective output projections were then averaged to provide a single (dithered) projection corresponding to the MNIST image projected with respect to the given *projection* (CIFAR) *image*. Both parallel dither during training of the projector DNNs and parallel dither during projections of the MNIST images are compared with the default (non-dithered) instances of the same paradigm.

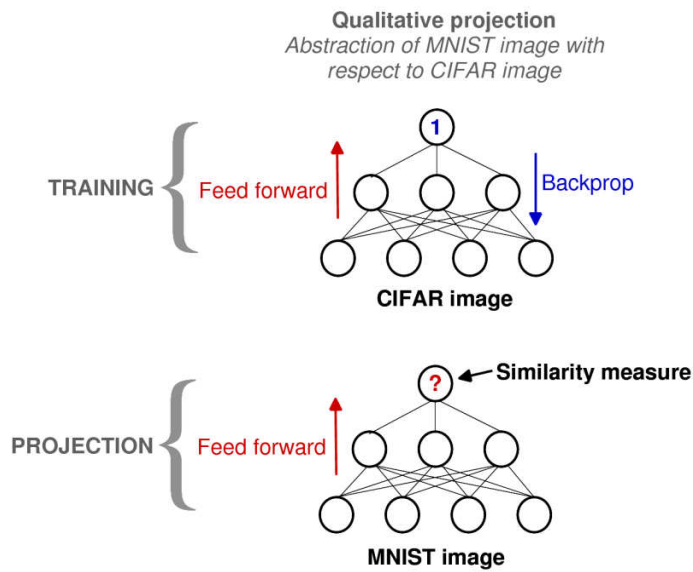

**Fig. 2. Q***ualitative projection* **via DNN – schematic diagram.** Each *projector DNN* was first trained (upper diagram) to abstract a '1' (magnitude) from a given CIFAR image. The trained *projector DNN* (lower diagram) was then used to abstract an MNIST image *with respect to* the CIFAR image. This provided a similarity measure (between the MNIST image and the CIFAR image).

*Qualitative projection classifier*. The 100 *projector DNN*s were used to obtain a multi-dimensional *qualitative projection* (of dimension 100) for each 1000 training and 10,000 test images from the MNIST dataset. For each MNIST image, projected with respect to each of the CIFAR images, this provided a vector of dimension 100. Using these qualitative projections of the training and test MNIST images, a *classifier DNN* of size 100x100x10 was instantiated, with the 100-way input layer corresponding to the qualitative projection vector and 10-way softmax output layer. Biased-sigmoid activation functions were used (with zero bias in the output layer). For each possible configuration of the qualitative projection paradigm (biased sigmoids / non-biased sigmoids, dithered / not-dithered), a classifier DNN was trained, using the 1000 qualitative-projected training images, for 150 full-sweep iterations. After each iteration of training, test error was computed with respect to the 10,000 test images. Each instance of the classifier DNN was trained from the exact same random starting weights. A learning rate of 1 was used in each case. Training was performed using typical non-batch stochastic gradient descent. Dither [3-5] was not used to train the classifier for reasons that are somewhat beyond the scope of this article.

### III. RESULTS

Fig. 3 plots test error rate (over 10,000 *qualitatively projected* test images) for the *classifier DNN* trained on the *qualitative projections* taken from the MNIST images with respect to the CIFAR images. The blue trace plots the evolution of test error with training for the *classifier DNN* trained and tested with the projections obtained without any regularisation to enforce demodulation (no biased-sigmoid and no dither). The results are around chance, indicating that the process of *qualitative projection* fundamentally failed to capture any meaningful qualities. Presumably, this is due to the capacity of the un-regularised *projector DNN*s for simply learning the simplest (arbitrary) path to obtaining a '1' in the output layer.

The green trace of Fig. 3 plots the same results for the *projector DNN*s which featured the biased-sigmoid activation function, which is optimised for demodulation (see [2]). In this case, the *qualitative projection* process has been successful and the *classifier DNN* has been able to generalise the *qualitative projection* features sufficiently to obtain an error rate near to 50%. Presumably, then, this improvement is the result of the biased-sigmoid activation function acting as regulariser to enforce demodulation within the *projector DNN*s.

The red trace plots the same results for the *projector DNN*s both featuring biased-sigmoids and parallel dither during training and projection. Thus, this instance of the projection process featured regularisation to both enforce demodulation and to suppress 'decoy features'. As a result, the error rate approaches 30%.

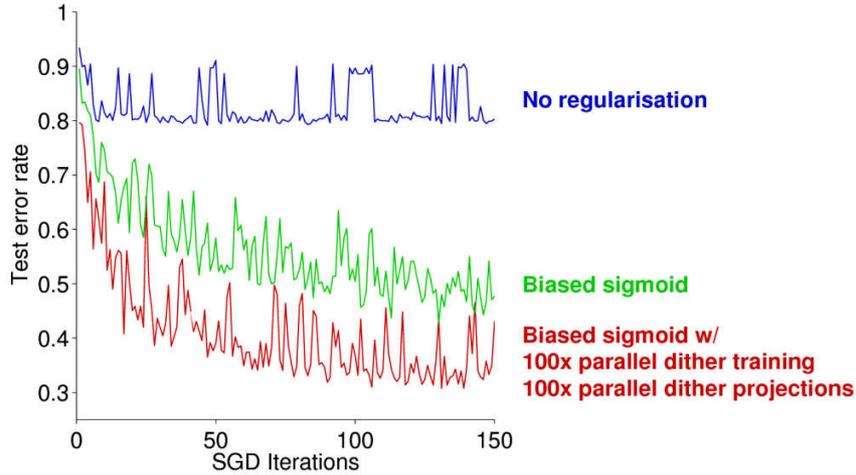

**Fig. 3. Qualitative projection of MNIST digits with respect to CIFAR images.** *Classifier DNN* (trained on the qualitative projections of MNIST images) test error computed over the 10,000 test images of MNIST as a function of training iterations for each of the three possible *projector DNN* configurations; **blue**: training and projections of *projector DNN*s awithout regularisation for abstraction (no biased sigmod, no dither), **green**: training and projections of *projector DNN*s regularised for abstraction with biased sigmoid (biased for optimum demodulation [2]) and **red**: training and projections of *projector DNN*s regularised for abstraction with biased sigmoid and *parallel dither* [3-5] applied to both training and projection.

## IV. DISCUSSION AND CONCLUSION

In this article, we have extended the discrete signal processing interpretation of DNN [1-5] with a refined definition of abstraction. This refined definition provides access to the qualitative nature of abstraction, and in particular to the process by which abstraction is *qualified*. We have illustrated the traction afforded by this by describing a process of *qualitative projection*. We have obtained abstractions of MNIST hand-written digits with respect to arbitrary images from the CIFAR dataset and we have demonstrated that these projections may be used to train a digit classifier without resorting to the native feature space of the digit images. By contrasting performance with varying degrees of regularisation to enforce abstraction, we have illustrated the essential nature of abstraction and we have captured this with the concept of *qualitative projection*.

## V. APPENDIX: HOW TO IDENTIFY QUALITATIVE TERMS IN EQUATIONS

*Qualification* exists because it is useful to make comparisons. In this article, we have argued that *qualification* is a *process*. We have argued that *qualification* is the defining process of *abstraction* and we have demonstrated empirically that our rigorous definition of *Qualitative Projection* is inherently useful for comparison (classification). We have provided a rigorous, signal processing definition of abstraction: *abstraction is the process whereby the output of a linear filter is demodulated*. Thus, the abstraction of an input is *qualified by* the linear filter.

Our definition of *Qualitative Projection* involves the identification of the two sequential (directional) signal processing operations - linear filtering and nonlinear demodulation. Thus, we may use the same identification process to identify *Qualitative Projection* in other equations (than those of DNN). In this brief appendix, we use our definition of *Qualitative Projection* to identify *Qualitative Projections* occuring in physics. By a method of 'like terms', we identify *volume*, *mass*, *energy* and *gravity* as *Qualitative Projections* and we briefly discuss the implications of directionality in this logic.

*Body as Qualifier.* Central to most of physics is the notion of a *body*. A *body* is defined by a contiguous boundary in space (usually 3D space). The matter contained within this boundary is collectively objectified by the boundary. In the simplest instantiation, a 1D body spans (is bounded by) two points (e.g., on the line of space). Given the coordinates of the two points $(x_1, x_2)$, the length ($L$) of the body is computed by demodulating the derivative:

$$L = abs(x_2 - x_1) \qquad (1)$$

In signal processing terms, this can be rewritten as a linear filtering operation such that

$$L = abs(a_1 x_1 + a_2 x_2) \qquad (2)$$

where the filter coefficients $a_1$ and $a_2$ are equal to -1 and 1 respectively (a high-pass filter). Thus, we identify a linear filter, and in the *abs* operation we identify demodulation. This may be restated using a more familiar (almost Pythagorean) nonlinearity:

$$L^2 = (a_1 x_1 + a_2 x_2)^2 \qquad (3)$$

Thus, we identify length ($L$) as a *Qualitative Projection* – *the abstraction of length of the body from the coordinates of the body is qualified by the filter by which we define the boundaries of the body*.

*Volume.* To generalise this, let us consider the 3D *volume* of a sphere. The volume ($V$) of a sphere is defined with respect to its radius ($r$) as:

$$V = \tfrac{4}{3}\pi r^3 \qquad (4)$$

The radius ($r$) of a sphere is a *magnitude* and may be obtained for a sphere with center $(x_1, y_1, z_1)$ by:

$$r^2 = (x - x_1)^2 + (y - y_1)^2 + (z - z_1)^2 \quad (5)$$

This may be restated in terms of the implicit linear filters:

$$r^2 = (a_1 x + a_2 x_1)^2 + (b_1 y + b_2 y_1)^2 + (c_1 z + c_2 z_1)^2 \quad (6)$$

Where the boundary-defining filter coefficients $(a_1, a_2)$, $(b_1, b_2)$ and $(c_1, c_2)$ are equal to (1,-1) respectively (high-pass filters in 3D). Thus, we identify both the linear filter and the demodulation operations (the square terms are demodulators). Thus, we identify that *volume* is a *Qualitative Projection* – the abstraction of volume of the body from the coordinates of the body is qualified by the filter by which we define the boundaries of the body.

*Mass, Energy.* Mass is defined as density multiplied with volume. Hence, mass is a scaled version of volume. From here, Einstein's equation [10]:

$$E = mc^2 \quad (7)$$

tells us that energy (*E*) is obtained with a further rescaling (via the speed of light, *c*, squared). Thus, *mass* and *energy* come from rescaling of volume and hence are identified as *Qualitative Projections*.

*Gravity.* Let us consider two *bodies* with centers $(x_1, y_1, z_1)$ and $(x_2, y_2, z_2)$, and defined with spherical boundaries whose *i*th radii (*r*) are abstracted using Eq. 6 and whose volumes (*V*) are computed:

$$V_i = \tfrac{4}{3}\pi r_i^3 \quad (8)$$

The respective masses ($m_1$ and $m_2$) are computed according to the respective densities ($D_1$ and $D_2$):

$$m_i = V_i D_i \quad (9)$$

and Newton's Gravity [11] (*F*) is defined:

$$F = G \frac{m_1 m_2}{r_G^2} \quad (10)$$

Where *G* is the gravitational constant, where the gravitational radius (or, gravitational *field*) ($r_G$) is defined according to derivative of the centers of the (spherical) bodies (in the implicit signal processing filter form):

$$r^2 = (a_1 x + a_2 x_1)^2 + (b_1 y + b_2 y_1)^2 + (c_1 z + c_2 z_1)^2 \quad (11)$$

Where, as previously, the filter coefficients are (-1,1) respectively. Thus, we identify gravity as a *Qualitative Projection* – the abstraction of gravity of the two bodies from the center coordinates and masses of the two bodies is first qualified by the filters by which we define the boundaries of the two bodies and qualified second by the filters by which we define the boundaries of the gravitational field between the centers of the two bodies.

Note that this leads to our recognition of the gravitational field as being indistinguishable from a *body* (defined by boundaries in space). Both *Qualitative Projections* (first of volume/mass, then of gravity) are qualified by the filters representing the *bodies*. Thus, as a *body*, the definition of the gravitational field is as arbitrary a qualification as the definitions of the bodies it 'acts upon'. Thus, neither the field (*body*) nor the bodies are recognised by the physical universe – they are both merely necessary qualifications for the abstraction of gravity. Thus, the abstraction of gravity enables us to *qualitatively* compare the two bodies.

*Implications for interpretation.* A consequence of this identification process is that it becomes clear that volume, mass, energy and other qualitative projections (such as *probability*) are *non-physical*. This is to say that they are abstractions *qualified by* arbitrary definitions of *body*. This tells us that the physical universe does not recognise the notion or definition of *body*. Thus, terms such as volume, mass, energy and gravity (and probability) are arbitrarily useful for comparison but do not physically exist and may not be observed in the universe.

Our process of qualification is strictly directional. I.e., there are infinite arrangements of a *body* which satisfy a given equation of volume, mass, etc. Thus, *we cannot use the abstractions (e.g., gravity) in order to define the body*. Or, to restate, we cannot use the abstractions to violate the qualifications which inherently qualify the abstractions. This is the definition of a paradox. It is therefore not surprising that most of the paradoxes of physics originate in propositions featuring gravity or probability (or, worse yet, both).

ACKNOWLEDGMENT

AJRS did this work on the weekends and was supported by his wife and children.